\documentclass[preprint,12pt]{elsarticle}

\usepackage{amssymb}
\usepackage{framed,multirow}

\usepackage{latexsym}
\usepackage{textcomp,booktabs}
\usepackage{amssymb}
\usepackage{pifont}
\newcommand{\cmark}{\ding{51}}%
\newcommand{\xmark}{\ding{55}}%

\usepackage{url}
\usepackage{cite}
\usepackage{natbib}
\usepackage{xcolor}
\definecolor{newcolor}{rgb}{.8,.349,.1}
\usepackage{marvosym}

\usepackage{hyperref} 
\hypersetup{hidelinks, colorlinks=true, linkcolor=red, citecolor=blue, urlcolor=magenta}

\journal{Pattern Recognition}

\begin{document}

\begin{frontmatter}




\title{ Semantic-Aware Frame-Event Fusion based Pattern Recognition via Large Vision-Language Models}






\author{ Dong Li\textsuperscript{1}, Jiandong Jin\textsuperscript{2}, Yuhao Zhang\textsuperscript{1}, Yanlin Zhong\textsuperscript{1}, \\ 
        Yaoyang Wu\textsuperscript{1}, Lan Chen\textsuperscript{3}\textsuperscript{\Letter}, Xiao Wang\textsuperscript{1}\textsuperscript{\Letter}, Bin Luo\textsuperscript{1}} 
\address{
1. School of Computer Science and Technology, Anhui University, Hefei 230601, China. \\
2. School of Artificial Intelligence, Anhui University, Hefei 230601, China. \\
3. School of Electronic and Information Engineering, Anhui University, Hefei 230601, China. \\
}


\begin{abstract}
Pattern recognition through the fusion of RGB frames and Event streams has emerged as a novel research area in recent years. Current methods typically employ backbone networks to individually extract the features of RGB frames and event streams, and subsequently fuse these features for pattern recognition. However, we posit that these methods may suffer from two key issues: 
1). They attempt to directly learn a mapping from the input vision modality to the semantic labels. This approach often leads to sub-optimal results due to the disparity between the input and semantic labels; 
2). They utilize small-scale backbone networks for the extraction of RGB and Event input features, thus these models fail to harness the recent performance advancements of large-scale visual-language models. 
In this study, we introduce a novel pattern recognition framework that consolidates the semantic labels, RGB frames, and event streams, leveraging pre-trained large-scale vision-language models. Specifically, given the input RGB frames, event streams, and all the predefined semantic labels, we employ a pre-trained large-scale vision model (CLIP vision encoder) to extract the RGB and event features. To handle the semantic labels, we initially convert them into language descriptions through prompt engineering, and then obtain the semantic features using the pre-trained large-scale language model (CLIP text encoder). Subsequently, we integrate the RGB/Event features and semantic features using multimodal Transformer networks. The resulting frame and event tokens are further amplified using self-attention layers. Concurrently, we propose to enhance the interactions between text tokens and RGB/Event tokens via cross-attention. Finally, we consolidate all three modalities using self-attention and feed-forward layers for recognition. Comprehensive experiments on the HARDVS and PokerEvent datasets fully substantiate the efficacy of our proposed SAFE model. 
The source code will be made available at \textcolor{magenta}{\url{https://github.com/Event-AHU/SAFE_LargeVLM}}.   
\end{abstract}


\end{frontmatter}


\section{Introduction}  
RGB camera-based pattern recognition is a foundation problem that has been widely explored and further boosted by deep learning. However, the performance may encounter bottlenecks due to the incomplete input information of RGB cameras in some scenarios. The influence factors may include low illumination, high-speed motion, etc. In order to solve these problems, some new sensors have been developed, among which pattern recognition based on Event cameras has attracted more and more attention~\citep{wang2023sstformer, jiang2023PVGraph, wang2022hardvs}. Different from RGB cameras which output video frames synchronously by recording light intensity in the scene with a fixed frame rate, Event camera (also termed Dynamic Vision Sensor, DVS) is a bio-inspired newly developed sensor that captures the variation of light intensity asynchronously and emits an event/spike when the change exceeds a threshold. The event point corresponding to the increase/decrease of light intensity is termed an ON/OFF event, respectively. Usually, a quadruple is utilized to represent the event $[x, y, t, p]$, where $(x, y)$ is the spatial coordinates of the event, $t$ and $p$ denote the time stamp and the polarity, respectively. As validated in many works~\citep{gallego2020eventsurvey}, the Event cameras perform better than RGB cameras on the high dynamic range, energy consumption, low illumination, fast motion, etc.

As the event signals are weak in motionless scenarios and also can't reflect the detailed texture or color information of target objects, it is still necessary to incorporate the RGB cameras in these cases for more accurate pattern recognition. Similar views can also be found in existing event-based tasks~\citep{yuan2023bottleneck}. There are already some works considering fusing these two modalities for more robust recognition. To be specific, 
Yuan et al.~\citep{yuan2023bottleneck} propose a dual-stream framework (Transformer and Structured Graph Neural Network) for event representation, extraction, and fusion by considering event images and voxels. 
Wang et al.~\citep{wang2023sstformer} develop a memory-support Transformer and spiking neural networks for RGB frames and event streams encoding, respectively. Also, the multi-modal bottleneck fusion module is introduced for RGB-Event feature aggregation.

\begin{figure*}
    \centering
    \includegraphics[width=\linewidth]{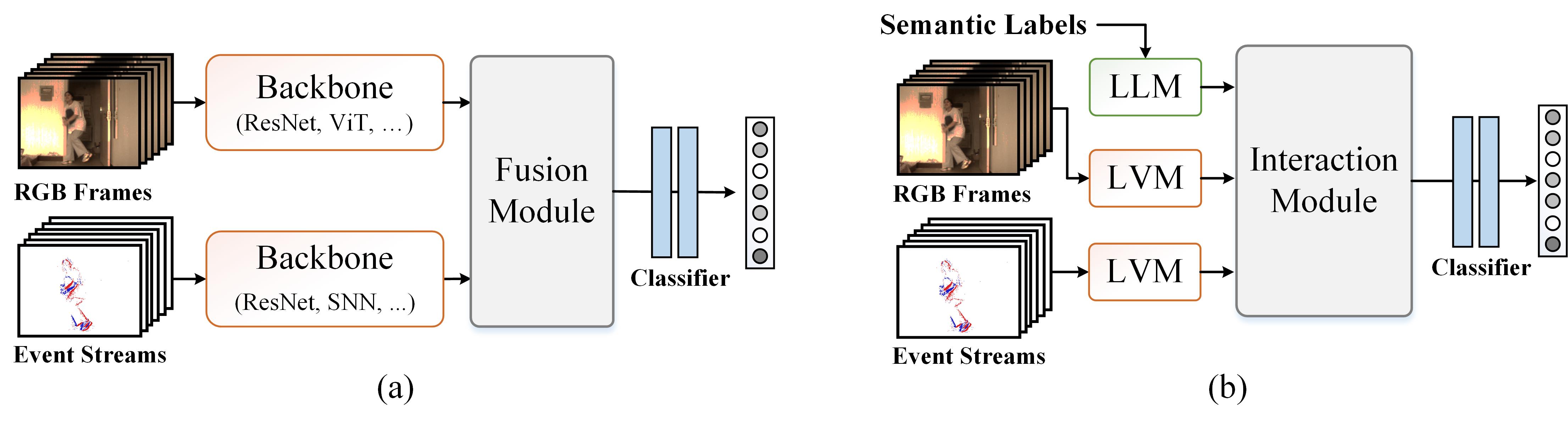}
    \caption{Comparison between (a) Standard RGB-Event Fusion Framework and (b) Large Model guided Semantic-aware RGB-Event Recognition. }
    \label{fig:firstIMG}
\end{figure*}

Although good performance has already been achieved on current benchmark datasets like PokerEvent~\citep{wang2023sstformer} and HARDVS~\citep{wang2022hardvs}, however, their results may still be limited due to the following issues: 
1). Existing works usually adopt CNN/Transformer pre-trained on classification datasets, e.g., ImageNet~\citep{deng2009imagenet}, but this approach does not enjoy the advantages and benefits of large models pre-trained on large amounts of data. More and more works~\citep{wang2023MMPTMs} demonstrate that the pre-trained big models could improve the generalization of downstream tasks significantly. 
2). Current works usually treat the RGB-Event based recognition task as a mapping from multi-modal inputs to one-hot representations. However, the semantic information of the target category is ignored, thus, there will be significant gaps between the RGB/Event data with the target category. An illustration of traditional recognition frameworks can be found in Fig.~\ref{fig:firstIMG} (a). 
Recently, multi-modal pre-trained models (MM-PTMs) have been proposed one after another~\citep{wang2023MMPTMs}, such as CLIP~\citep{radford2021learning} and ALIGN~\citep{jia2021scaling}. The generalization ability of such multi-modal pre-trained models outperforms traditional small-scale deep models significantly. Therefore, it is natural to raise the following question: \emph{``Can we bridge the gap between the RGB/Event data and semantic information of target categories and achieve more accurate pattern recognition based on pre-trained large multi-modal models?"}

Inspired by the aforementioned observations and reflections, in this paper, we propose a novel RGB-Event recognition framework by fusing the RGB frames, Event streams, and semantic category information based on large multi-modal pre-trained models, as shown in Fig.~\ref{fig:firstIMG} (b). To be specific, our framework contains three main modules, including the input embedding module, multi-modal fusion and interaction module, and classification head. We take the RGB frame clips, event streams, and the semantic labels of all categories as the inputs. For the RGB and Event data, we adopt the pre-trained Large Vision Model (LVM) for the feature extraction. For the semantic labels, we first expand and transform them into corresponding language descriptions using a prompt template. For instance, the human activity ``\textit{Pouring water}" can be transformed into a sentence ``\textit{The action of the human is \underline{pouring water} }". Then, a Large Language Model (LLM) is adopted for the feature embedding of the obtained sentence. We adopt a multi-modal Transformer to fuse the RGB and language embedding,  and Event and language embedding, respectively. Then, the output RGB and Event tokens are concatenated and fed into a self-attention layer. After that, two cross-attention layers are proposed to further boost the interactions between the vision and language features. Finally, we fuse these three features using self-attention and map them into target categories using feed-forward layers. An overview of our proposed RGB-Event pattern recognition framework is illustrated in Fig.~\ref{fig:framework}.

To sum up, the contributions of this work can be summarized as the following three aspects: 

$\bullet$ We rethink the RGB-Event based pattern recognition task and re-formulate it as a vision-language fusion problem that connects the vision and language modality perfectly. 

$\bullet$ We propose a new Semantic-Aware Frame-Event fusion based pattern recognition framework, termed SAFE, based on a large vision-language pre-trained model, which fuses the RGB frame, event data, and semantic label set in a unified framework. 

$\bullet$ Extensive experiments on two RGB-Event based recognition benchmark datasets, including PokerEvent~\citep{wang2023sstformer} and HARDVS~\citep{wang2022hardvs} datasets, fully validated the effectiveness of our proposed framework.

The organization of this paper can be summarized as follows: 
We first review the previous works that are most related to our framework in Section~\ref{relatedWorks}. 
Then, we focus on introducing our RGB-Event-Language fusion based framework in Section~\ref{Methodology}, including the overview, input encoding, feature enhancement module, and loss function. After that, we dive into the experimental results in Section~\ref{experiments} and will introduce the datasets and evaluation metrics used in this paper first. Then, we will describe the implementation details, comparison on public benchmark datasets with other state-of-the-art algorithms, ablation study, and parameter analysis. We also give some visualizations and limitation analysis in Section~\ref{visualizations} and \ref{limitations}. Finally, we summarize this paper and give a discussion on future works in Section~\ref{conclusions}.

\section{Related Works} \label{relatedWorks}
In this section, we will survey and discuss some significant methodologies that are pertinent to our work, particularly focusing on RGB-based Recognition, Event-based Recognition, and Vision-Language Fusion. More related works about Event cameras~\footnote{\url{https://github.com/Event-AHU/Event_Camera_in_Top_Conference}} and vision-language pre-trained models can be found in the following surveys~\citep{gallego2020eventsurvey, wang2023MMPTMs}.

\subsection{RGB-based Recognition} 
The crux of RGB-based recognition research hinges on the effective extraction and utilization of spatial and spatiotemporal features. 
Convolutional neural networks (CNN) \citep{r1}, \citep{r2}, \citep{r3}, \citep{r4}, founded on deep learning theory, represent one of the earliest methods used for processing RGB data. 
Karen et al. \citep{5550411545ce0a409eb386d9} propose a two-stream neural network, which is divided into Spatial Stream ConvNet and Temporal Stream ConvNet. 
Ji et al.\citep{r6} expanded traditional CNN to 3DCNN, incorporating temporal information and performing feature computation on both the temporal and spatial dimensions of video data. 
Chen et al. \citep{7926606} propose a Semi-Coupled Two-stream Fusion ConvNet to help train the network with low-resolution videos. 
Recurrent neural networks (RNN) \citep{r7}, \citep{r8}, \citep{r9} and Long short-term memory (LSTM) \citep{r10}, \citep{r11} have also been extensively employed in the processing of RGB data. 
RNN has seen rapid development, especially in managing temporal sequence data. 
Sak et al. \citep{r12} propose an LSTM-based method to address the vanishing gradient problem in RNN models, effectively capturing the dynamic changes in video sequences. 
Lakhal M.I. et al. \citep{5c5104cde1cd8eb4d3576e78} propose a method of using residual stacked recurrent neural networks (Res-RNN) for action recognition. 
Li et al. \citep{5b076eb4da5629516ce73bc9} propose a Video-LSTM for end-to-end sequence learning of action videos. 
Graph neural networks (GNN) and Transformers have also been utilized in RGB-based Recognition. 
For instance, Han et al. \citep{r13} used GNN to process spatial structural information in images, while Vaswani et al. \citep{r14} introduced the Transformer model, which outperforms and supersedes RNN and CNN with attention mechanisms, demonstrating higher parallelism and effectively handling the spatio-temporal features of RGB data. 
Mazzia V. et al. \citep{60dfd5b491e01129379b3837}  propose the Action Transformer (AcT) model with the application of the Transformer encoder. 
Although these models work well on simple videos, however, their performance may still be poor in challenging scenarios due to the utilization of RGB frames.

\subsection{Event-based Recognition} 
Compared to traditional RGB-based recognition, Event-based recognition places greater emphasis on recognition algorithms for event cameras. The prevailing approaches for event-based recognition can be divided into the following categories: CNN-based~\citep{8953966}, GNN-based~\citep{xie2022vmv}, \citep{r20}, and SNN-based~\citep{r21}, \citep{r22}. 
Given the sensitivity of event cameras to scene edges and motion directions, Time Surfaces (TSs)~\citep{r15}, \citep{r16} have been employed in numerous tasks related to motion analysis and shape recognition. 
CNNs also have widespread applications in this domain. Zhu et al. \citep{r17} proposed a CNN-based event camera that uses TSs as input to compute optical flow. 
For point cloud representation, Wang et al. \citep{r23} treated event streams as a series of 3D points in spacetime, termed as Spacetime Event Cloud and utilized the PointNet \citep{r24}, which takes point clouds as input and outputs class labels for the entire input or segment/portion labels for each input point. 
Xie et al. \citep{xie2022vmv} introduced VMV-GCN, a voxel-based geometric learning model designed to integrate multi-view volumetric data. 
Li et al. \citep{r25} proposed the Event Transformer to directly process event sequences in their native vectorized tensor format. 
Wang et al. \citep{wang2023sstformer} proposed a hybrid SNN-ANN framework, termed SSTFormer, which bridges the gap between SNN and (Memory Support Transformer) MST. 
Wang et al. \citep{8953966}  proposed a CNN-based gait recognition method, which effectively removes noise via motion consistency. 
Wu et al. \citep{wang2022hardvs} first transform the event flow into images, then, predict and combine the human pose with event images for HAR. 
Xing et al. \citep{5fdb2d6cd4150a363c998875} design a spiking convolutional recurrent neural network (SCRNN) architecture for event-based sequences.
Concurrently, SNN, an emerging neural network model, has also showcased its potential in Event-based Recognition. 
Lee et al. \citep{r26} proposed an SNN-based event camera recognition algorithm that utilizes computationally inspired supervised learning types, such as backpropagation, in deep networks to effectively implement spike-based deep convolutional networks. 
In contrast to previous works, this paper designs a Transformer-based fusion method for RGB-Event recognition based on a pre-trained large vision-language model. It bridges the modality gaps between vision and semantic labels and achieves higher recognition performance.

\subsection{Vision-Language Fusion} 
In recent years, Vision-Language Fusion has surfaced as a significant research direction. 
The crux of this field lies in the effective integration of visual and language information. 
One intuitive approach to fuse modalities is through straightforward operations such as weighted addition or concatenation \citep{r27}, \citep{r28}, \citep{r29} to amalgamate features from different modalities. Contemporary mainstream multi-modal models utilize Transformer encoders for deep cross-modal fusion. By concatenating different modality sequences, this design allows for unrestricted fusion of cross-modal information. 
Some noteworthy research includes Li et al.\citep{r29}, who proposed a method called VisualBERT for fusing visual and language information. VisualBERT comprises a stack of Transformer layers that implicitly align the elements of input text with the relevant regions in the input image through self-attention. This method effectively manages and understands image and text-related information. 
Chen et al.\citep{r30} introduced UNITER, a universal image-text representation method, which accomplishes effective processing and fusion of visual and language information through large-scale pre-training on four image-text datasets. 
Pre-trained multi-modal big model CLIP \citep{radford2021learning} have shown great potential in learning representations that are transferable across a wide range of downstream tasks. 
Zhou et al. \citep{613192dd5244ab9dcb9e5af8} proposed Context Optimization (CoOp), a simple approach for adapting CLIP-like vision-language models for downstream image recognition. 
Jia et al. \citep{jia2021scaling} proposed a method for extending visual and vision-language representation learning, which addressed the challenges posed by noisy text supervision. 
Kim et al. \citep{60211a1e91e0113bfb1dc6a2} proposed a minimal VLP model, Vision-and-Language Transformer (ViLT), which removes the complexity of deep visual embeddings and region supervision. 
Cheng et al. \citep{62cec7b65aee126c0f5794bc}  proposed a novel visual-textual baseline (VTB) for PAR to explore the textual semantic correlations from attribute annotations by pre-trained textual encoders instead of human definitions. 
Inspired by these works, in this paper, we propose a novel RGB-Event-Language fusion framework for the RGB-Event based pattern recognition.

\section{Methodology} \label{Methodology} 
In this section, we will first give an overview of our proposed framework in sub-section~\ref{overview}. After that, we will dive into the details of our approach, with a focus on the input encoding, feature enhancement module, and loss function in sub-section~\ref{inputencoding}, \ref{featEnhancement}, and \ref{lossFunction}, respectively.

\begin{figure*}
    \centering
    \includegraphics[width=\linewidth]{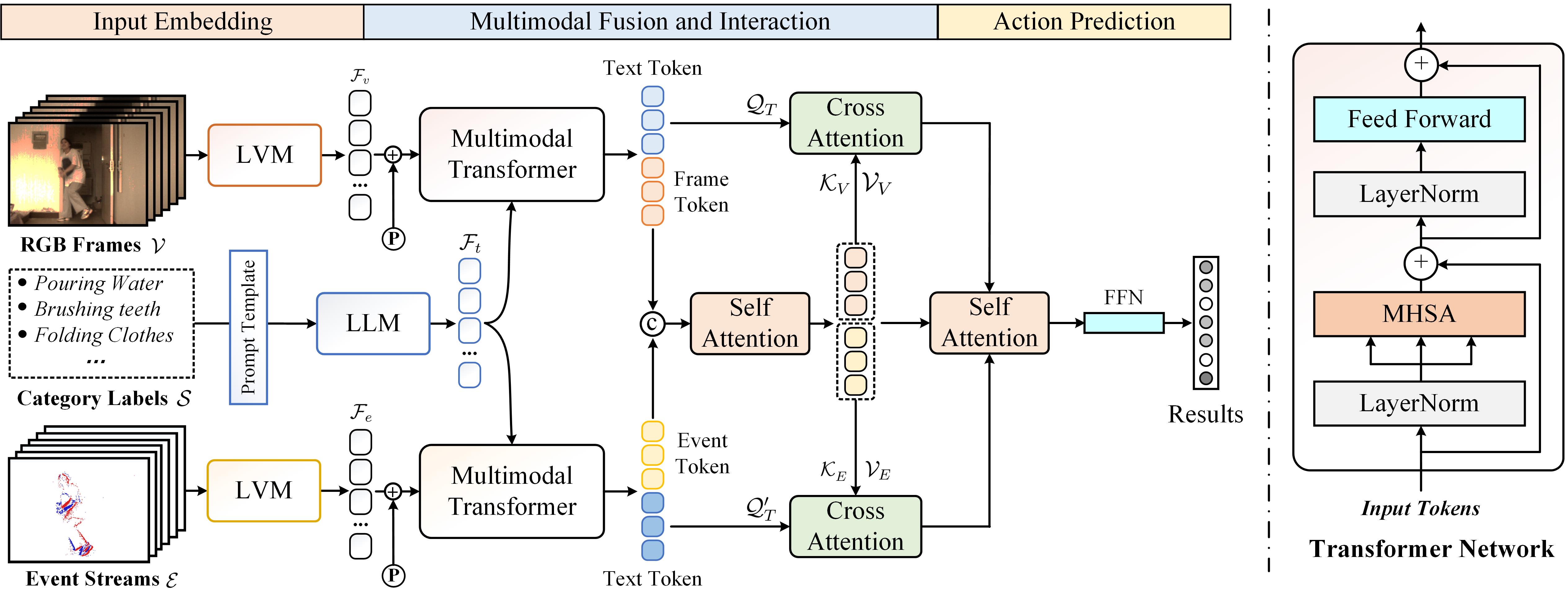}
    \caption{An overview of our proposed Semantic-Aware Frame-Event fusion based pattern recognition framework, termed SAFE.} 
    \label{fig:framework}
\end{figure*}

\subsection{Overview} \label{overview} 
As illustrated in Fig.~\ref{fig:framework}, our proposed framework takes the RGB frames, Event streams, and semantic category information as the input. We adopt the pre-trained multi-modal models to process these inputs respectively. To be specific, the Large Vision Models (LVM) are adopted to encode the RGB frames and event streams. For the semantic category information, we first expand the semantic labels into language descriptions based on the prompt template. Then, a Large Language Model (LLM) is utilized to achieve the language embedding. After that, we fuse the RGB and language features, event and language features using multi-modal Transformers, separately. The output frame and event tokens are enhanced using a self-attention scheme and then fused with text tokens via cross-attention. Finally, we fuse these three features using self-attention and map them into category labels using feed-forward layers. In the following subsequent sections, we will focus on each detailed module of our proposed framework.

\subsection{Input Encoding} \label{inputencoding} 
Given the video frames $\mathcal{V} = \{v_1, v_2, ..., v_N\}$ and event streams $ \mathcal{E} = \{e_1, e_2, ..., e_M\}$, and the defined semantic category labels in each dataset $\mathcal{S} = \{s_1, s_2, ..., s_L\}$, where $N, M$ and $L$ denotes the number of video frames and event points, and semantic labels. Note that, each point in the event stream can be represented as $e_i = [x, y, t, p]$, where $(x, y)$ denotes the spatial coordinates, $t$ and $p$ represent the timestamp and polarity.   
We first transform the event streams $\mathcal{E}$ into event images according to the timestamp of video frames. To better obtain the feature representations of these inputs, in this work, we adopt the pre-trained CLIP model~\citep{radford2021learning} which contains the Large Language Model (LLM) and the Large Vision Model (LVM). We first pre-process the input frames by resizing them into a fixed resolution $224 \times 224$. The ViT-B/16 based CLIP model is adopted to achieve a better trade-off between accuracy and efficiency. Therefore, we can get a set of visual tokens  $\mathcal{T}_v^{N \times 197 \times 768}$, and event tokens  $\mathcal{T}_e^{N \times 197 \times 768}$, where $197$ and $768$ represent the number of tokens and the dimension of each token, respectively. 
After feeding the RGB and Event tokens into two LVMs, we can get its corresponding enhanced features: 
\begin{equation}
    \mathcal{F}_v =  {LVM}(\mathcal{T}_v),~~~~\mathcal{F}_e =  {LVM}(\mathcal{T}_e).
\end{equation}

To better utilize the semantic category labels $\mathcal{S} = \{s_1, s_2, ..., s_L\}$, in this work, we transform the semantic labels into language descriptions using prompt engineering. For example, we design a template \emph{``The action of the human is \underline{~~~~~~~~}"} and fill in the blanket using the corresponding ground truth semantic label, i.e., \emph{``The action of the human is \underline{eating an apple}"}. Then, we adopt the CLIP text encoder which is a large language model to get the textual tokens $\mathcal{F}_t = \{t_1, t_2, ..., t_L\}$.

\subsection{Feature Enhancement Module} \label{featEnhancement} 
In this work, we concatenate the video tokens and textual tokens as the multiple inputs $[\mathcal{F}_v, \mathcal{F}_t]$ and feed them into the multi-modal Transformer to capture the interactions between the visual and textual modalities. Similarly, we concatenate the event tokens $\mathcal{F}_e$ and textual tokens $\mathcal{F}_t$ as $[\mathcal{F}_e, \mathcal{F}_t]$, and then feed them into another multi-modal Transformer. By leveraging the CLIP model's visual and textual encoders, we can effectively capture the visual and semantic information of the input data, facilitating multi-modal fusion and interaction. 
The core operation of a multi-modal Transformer network is the self-attention layer which takes the visible/event and language tokens as the input. For example, the visual and language tokens $[\mathcal{F}_v, \mathcal{F}_t]$ are transformed into query $\mathcal{Q}_{VT}$, key $\mathcal{K}_{VT}$, and value $\mathcal{V}_{VT}$, respectively. Then, the self-attention can be formulated as: 
\begin{equation}
    \label{selfattention} 
    SelfAtt_{VT} = Softmax(\frac{\mathcal{Q}_{VT} \mathcal{K}_{VT} ^T}{\sqrt{k}}) \mathcal{V}_{VT},     
\end{equation}
where $Softmax$ denotes the Softmax operation, $k$ is the dimension of feature vector. 
Similarly, the self-attention in the Event-Text Multi-modal Transformer which takes the transformed $\mathcal{Q}_{ET}$, $\mathcal{K}_{ET}$, and $\mathcal{V}_{ET}$ can be represented as: 
\begin{equation}
    \label{selfattention} 
    SelfAtt_{ET} = Softmax(\frac{\mathcal{Q}_{ET} \mathcal{K}_{ET} ^T}{\sqrt{k}}) \mathcal{V}_{ET}.    
\end{equation}

After we get the enhanced features using the multi-modal Transformer network, we split them into text tokens and visible/event tokens, respectively. Then, a self-attention scheme is adopted to fuse the visible and event tokens. The output will be fed into the cross-attention layers along with text tokens, as illustrated in Fig.~\ref{fig:framework}. In this procedure, the text tokens are used as the query feature $\mathcal{Q}_{T}$, and the visible ($\mathcal{K}_{V}$) /event tokens ($\mathcal{V}_{V}$) are utilized as the key and value features, i.e., 
\begin{equation}
    \label{selfattention} 
    CrossAtt_{VT} = Softmax(\frac{\mathcal{Q}_{T} \mathcal{K}_{V} ^T}{\sqrt{k}}) \mathcal{V}_{V}. 
\end{equation} 
Similar operations are conducted on the enhanced event-text tokens: 
\begin{equation}
    \label{selfattention} 
    CrossAtt_{ET} = Softmax(\frac{\mathcal{Q}'_{T} \mathcal{K}_{E} ^T}{\sqrt{k}}) \mathcal{V}_{E}. 
\end{equation} 
Finally, we concatenate all the feature tokens and feed them into a self-attention layer for multimodal fusion. Then, a classifier that contains one fully connected layer is used for pattern classification.

\subsection{Loss Function} \label{lossFunction}

In this study, we opted for the cross-entropy loss function as our loss function. The cross-entropy loss function is a widely employed loss function for classification tasks and is defined as follows:
\begin{equation}
L(y, \hat{y}) = -\sum_{i=1}^{N} y_i \log(\hat{y}_i)
\end{equation}
where \(y\) represents the ground true labels, \(\hat{y}\) represents the predicted labels, and \(N\) denotes the number of classes. The cross-entropy loss function measures the disparity between the true labels and the predicted labels.

\section{Experiments} \label{experiments}

\subsection{Datasets and Evaluation Metrics} 
In this section, we conduct extensive experiments on two event-based classification datasets, including HARDVS~\citep{wang2022hardvs} and PokerEvent~\citep{wang2023sstformer}. A brief introduction to the two datasets is given below and the detailed statistics can be found in Table~\ref{tab:datasets}. 

\begin{table}[!htp]
\center
\small   
\caption{The detailed statistics about the datasets used in our experiment.} 
\label{tab:datasets}
\resizebox{\columnwidth}{!}{ 
\begin{tabular}{l|cccc}
\hline 
Dataset             &\#Categories  &\#Samples  &\#Training Samples  &\#Testing Samples \\
\hline 
HARDVS      &300  &96908  &64522  &32386 \\
PokerEvent  &114  &24415  &16216  &8199 \\
\hline 
\end{tabular}
}
\end{table}

\noindent $\bullet$ \textbf{HARDVS Dataset.}\footnote{\url{https://github.com/Event-AHU/HARDVS}}~
This dataset is collected using a DVS346 event camera, which concentrates on recognizing human activities such as walking, running, and crouching. It consists of 300 classes and includes 96,908 RGB-Event samples. It is divided into training and testing subsets, with 64,522 and 32,386 samples respectively. 
 
\noindent $\bullet$ \textbf{PokerEvent Dataset.}\footnote{\url{https://github.com/Event-AHU/SSTFormer}}~
The PokerEvent dataset focuses on recognizing character patterns in poker cards. It consists of 114 classes and includes 24,415 RGB-Event samples recorded using a DVS346 event camera. The dataset is divided into training and testing subsets, with 16,216 and 8,199 samples, respectively.

For the evaluation of our and other compared recognition algorithms, we adopt the \textit{top-1 accuracy} as the evaluation metric.

\subsection{Implementation Details}  
In the training phase, we set the batch size to 16 and train the model for a total of 50 epochs. The learning rate is initially set to a fixed value of 8e-06, and the \textit{CosineLRScheduler} (with decay rate 0.1) is used to adaptively adjust the learning rate in the training procedure. AdamW optimizer~\citep{loshchilov2017decoupled} is adopted for the training of our network. For the HARDVS dataset, we conduct our experiments on four NVIDIA GeForce RTX 3090 GPUs which takes approximately 60 hours to complete. The training on the PokerEvent dataset takes about 30 hours using two NVIDIA GeForce RTX 3090 GPUs. More details can be found in our source code.

\subsection{Comparison on Public Benchmark Datasets}  
In this section, we will report our recognition results and compare them with other state-of-the-art algorithms on the PokerEvent and HARDVS datasets.

\noindent \textbf{Results on PokerEvent dataset.~}  
As shown in Table~\ref{poker_acc}, we compare multiple strong and classical recognition models, including C3D~\citep{tran2015learning}, TSM~\citep{lin2019tsm}, and TAM~\citep{liu2021tam}. In addition, we also compare with Transformer based models, such as V-SwinTransformer~\citep{liu2022video}, TimeSformer~\citep{bertasius2021space}, and MViT~\citep{li2022mvitv2}. We can find that our model achieves 57.64\% on this dataset which is significantly better than all the compared models. These experiments fully validated the effectiveness of our proposed SAFE model for RGB-Event based pattern recognition.


\begin{table}[!htp]
\center
\small   
\caption{Results on the PokerEvent dataset. } 
\label{poker_acc}
\begin{tabular}{l|c|c|c}
\hline 
\textbf{Algorithm} & \textbf{Source}      &\textbf{Backbone}  &\textbf{Precision}  \\
\hline
\textbf{ C3D } \citep{tran2015learning}              & ICCV-2015     &3D-CNN   &51.76     \\ 
\textbf{ TSM  } \citep{lin2019tsm}                   &  ICCV-2019   &ResNet-50   &55.43    \\ 
\textbf{ ACTION-Net } \citep{wang2021action}         &  CVPR-2021   &ResNet-50  &54.29    \\
\textbf{ TAM   } \citep{liu2021tam}                  &  ICCV-2021     &ResNet-50   &53.65    \\ 
\textbf{ V-SwinTrans   }\citep{liu2022video}         &  CVPR-2022     &Swin Transformer   &54.17    \\ 
\textbf{ TimeSformer   }\citep{bertasius2021space}   &  ICML-2021     &ViT-B/16  &55.69    \\ 
\textbf{ X3D   }\citep{feichtenhofer2020x3d}         &  CVPR-2020       &ResNet   &51.75     \\
\textbf{ MVIT   }\citep{li2022mvitv2}                &  CVPR-2022       &ViT   &55.02     \\
\textbf{ SSTformer   }\citep{wang2023sstformer}                &  arXiv-2023       &SNN-Former   &54.74     \\
\hline
\textbf{ SAFE (Ours)}                                      &--  &ViT-B/16   &57.64     \\ 
\hline
\end{tabular}
\end{table}

\noindent \textbf{Results on HARDVS dataset.~}
As shown in Table \ref{hardvs_acc}, our ViT-based model, SAFE, achieves a precision of 50.17\%. This result surpasses several prominent models such as ACTION-Net \citep{wang2021action} (46.85\%), SlowFast \citep{feichtenhofer2019slowfast} (46.54\%), and X3D \citep{feichtenhofer2020x3d} (47.38\%). Our model's performance is also competitive with ESTF \citep{wang2022hardvs} (49.93\%) and slightly falls short when compared to TimeSformer \citep{bertasius2021space} (51.57\%). These findings speak volumes about the effectiveness of our model, demonstrating that it can achieve comparable or even superior results than some state-of-the-art models. Our experiments on the HARDVS dataset serve as a valuable reference for other researchers and practitioners and provide encouraging insights into the potential of our model in this field.

\begin{table}
\center
\small   
\caption{Results on the HARDVS dataset. } 
\label{hardvs_acc}
\begin{tabular}{l|c|c|c}
\hline 
\textbf{Algorithm} & \textbf{Source}      &\textbf{Backbone}  &\textbf{Precision}  \\
\hline
\textbf{C3D}\citep{tran2015learning} & ICCV-2015     &\ 3D-CNN   &\ 50.88     \\
\textbf{ResNet18}\citep{he2016deep} & CVPR-2016     &\ ResNet18   &\ 49.20    \\
\textbf{ACTION-Net}\citep{wang2021action} & CVPR-2021     &\ ResNet-50   &\ 46.85     \\
\textbf{SlowFast}\citep{feichtenhofer2019slowfast} & ICCV-2019     &\ ResNet-50   &\ 46.54     \\
\textbf{R2Plus1D}\citep{tran2018closer} & CVPR-2018     &\ ResNet-34   &\ 49.06     \\
\textbf{TimeSformer} \citep{bertasius2021space}  &  ICML-2021     &\ ViT-B/16  &\ 51.57    \\ 
\textbf{X3D}\citep{feichtenhofer2020x3d}  &  CVPR-2020       &\ ResNet   &\ 47.38     \\
\textbf{ESTF}\citep{wang2022hardvs}  &  arXiv-2022       &\ ResNet-18   &\ 49.93     \\
\hline
\textbf{SAFE (Ours)}      &--  &\ ViT-B/16   &\ 50.17    \\ 
\hline
\end{tabular}
\end{table}

\subsection{Ablation Study}  

In this section, we conduct extensive experiments to further help the readers better understand the influence of each detailed setting on the final recognition results.

\noindent 
\textbf{Effects of Semantic Category Information.~}  
In this part, we will isolate each feature enhancement strategy for individual experiments to evaluate their impact on the final recognition performance. Specifically, we utilize the RGB and event images as the input of our network on the PokerEvent~\citep{wang2023sstformer} dataset, i.e., the semantic category information is removed in both the training and inference phases. As shown in Table~\ref{CAResults}, after removing the Semantic Category Information (SCI), we observed that the accuracy on the PokerEvent dataset decreased from $57.0\%$ to $56.6\%$. Based on these experiments, we can conclude that the proposed Semantic Category Information strategies indeed contribute to event-based recognition.

\noindent 
\textbf{Effects of Large Vision-Language Models.~}  
In our implementation, we adopt the pre-trained large vision-language model CLIP as the backbone for feature extraction, i.e., the Large-scale Visual Model (LVM) and large-scale language model (LLM). Specifically, when we replaced the large-scale visual model with the ViT model and trained it on the poker dataset, we observed a decrease in accuracy to $55.0\%$, as shown in Table~\ref{CAResults}. Based on this observation, we can conclude that the proposed strategy of using large-scale encoders contributes to RGB-Event based recognition.

\begin{table}
\center
\small     
\caption{ Component analysis of our proposed SAFE RGB-Event recognition model. 
``SCI" represents Semantic Category Information, ``LVM" represents Large Vision-Language Models, ``MT" represents Multi-modal Transformer, ``SA" represents Self-Attention, and ``CA" represents Cross-Attention.} 
\label{CAResults} 
\begin{tabular}{c|ccccc|c} 		
\hline 
\textbf{No.}  &\textbf{SCI} &\textbf{LVM}  &\textbf{MT}        &\textbf{SA}   &\textbf{CA}   &\textbf{Acc} \\
\hline 
1  &\cmark          &\cmark         &\cmark         &\cmark           &\cmark   &57.0       \\
2   &\xmark    &\cmark         &\cmark         &\cmark           &\cmark    &56.6       \\
3    &\cmark    &\xmark         & \cmark        &\cmark           &\cmark   & 55.0          \\
4    &\cmark   &\cmark    &\xmark         &\cmark           &\cmark   &55.5            \\
5    &\cmark    &\cmark   &\cmark   &\xmark         &\cmark 
  &56.2\\ 
6    &\cmark    &\cmark   &\cmark   & \cmark        &\xmark  &55.8\\ 
\hline
\end{tabular} 
\end{table}

\noindent 
\textbf{Effects of Multi-modal Transformer.~}  
In this work, we integrate the semantic category information into the RGB-Event recognition framework and introduce the Multimodal Fusion Transformer network to combine the visual and textual features. In order to validate the impact of the Multimodal Fusion Transformer on the entire network framework, we conducted an experiment where we did not fuse the visual and textual features through the Transformer. Instead, we directly concatenated the feature sequences while keeping the other parts unchanged. As the results reported in Table~\ref{CAResults}, we observed a decrease in accuracy to $55.5\%$ on the poker dataset, which further demonstrates the effectiveness of the Multimodal Fusion Transformer in the overall network framework.

\noindent 
\textbf{Effects of Self-Attention.~} 
Self-attention focuses on capturing the dependencies between different positions within a sequence. It allows each position to attend to all other positions in the sequence, capturing the importance or relevance of each position to the others. This mechanism enables the model to weigh the significance of different parts of the input sequence when making predictions or generating representations. Self-attention is particularly effective in capturing long-range dependencies and modeling relationships between different elements in the sequence. In our experiments, removing the self-attention module resulted in a decrease in accuracy to $56.2\%$ on the poker dataset, as shown in Table~\ref{CAResults}, indicating that the self-attention module also contributes to our framework.

\noindent 
\textbf{Effects of Cross-Attention.~} 
Cross-attention extends the self-attention mechanism to handle interactions between different sequences or modalities. Cross-attention is particularly useful in tasks involving multimodal data or information exchange between different modalities. In our experiments, removing the cross-attention module resulted in a decrease in accuracy to $55.8\%$ on the poker dataset, as shown in Table~\ref{CAResults}, further underscoring the positive impact of this module on our network framework.

\subsection{Parameter Analysis} 

\noindent 
\textbf{Results of Different Input Frames.~}  
In this study, we examined the impact of varying the number of input frames on the accuracy of our neural network model. Specifically, we tested models trained on 1, 3, 5, and 7 input frames, and analyzed the resulting accuracy scores. The experimental results are shown in Table~\ref{frames}. Contrary to what might be expected, our results did not demonstrate a consistent improvement in the model's accuracy as the number of input frames increased. The model that was trained on 5 input frames achieved the highest accuracy score of 57.02, whereas the model trained with only 1 input frame reached the lowest score of 53.92. Interestingly, the model trained with 7 input frames did not perform as well as the one with 5 frames, achieving an accuracy of 56.37. 

One possible interpretation of this pattern is that, while more input frames do provide the model with a broader view of the target object or action, there could be a threshold beyond which adding more frames doesn't contribute to additional useful visual information and might introduce noise instead. In our experiment, it appears that the optimal balance was achieved with 5 input frames. Another possible explanation is that while additional input frames help the model capture the motion or dynamics of the object or action better, too many frames might introduce a level of dynamic complexity that interferes with the model's performance. This seems to be exhibited in our experiment when the model's accuracy reduced slightly with 7 input frames compared to 5. 

In summary, our findings suggest that the number of input frames is a crucial parameter to consider when training a neural network model for video analysis tasks. While moderately increasing the number of input frames can lead to improved accuracy, an excessive number might introduce noise and require more computational resources and longer training times without a corresponding increase in accuracy.

\begin{table}
\center
\small   
\caption{Results of Different Input Frames. } 
\label{frames}
\begin{tabular}{l|c|c|c|c}
\hline 
\textbf{\#Frames}             &\textbf{1}   &\textbf{3}  &\textbf{5}     &\textbf{7} \\
\hline
\textbf{Results}    &  53.92           & 56.12          &     57.02          &    56.37      \\ 
\hline
\end{tabular}
\end{table}

\noindent 
\textbf{Results of Different Prompts.~}  
In this study, we delved into the influence of varying prompts on the accuracy of our neural network model. We experimented with models trained on prompts encapsulating different aspects of the target object or action and scrutinized the accuracy scores that ensued. 

Table~\ref{prompts} encapsulates our findings. It was observed that the model's accuracy was contingent on the prompt deployed. Certain prompts lead to greater accuracy scores than others, thereby underlining the significance of prompt selection in shaping the model's performance. For instance, the prompt ``\textit{This is a picture about Picture of...}" led the model to achieve an accuracy of 55.90. The prompt ``\textit{The action in the picture is...}" resulted in a slightly better accuracy of 56.58. The description ``\textit{A photo of a...}" obtained an accuracy of 56.29. Interestingly, the prompt ``\textit{The content of the playing card is...}" led to the highest accuracy of 57.64. Moreover, when no prompt was given, the model still managed to achieve a commendable accuracy of 57.02. 

The fluctuations in accuracy scores can be attributed to the diverse information each prompt offers about the target object or action. In a nutshell, our findings underscore that the choice of prompt is a crucial parameter in training a neural network model for video analysis tasks. The effectiveness of different prompts can vary, depending on the types of objects or actions. Different lengths of prompts may require different amounts of computing resources. 


\begin{table}
\center
\small   
\caption{Results of Different Prompts.  } 
\label{prompts}
\begin{tabular}{l|c|c}
\hline 
\textbf{Prompts}       &\textbf{Backbone}  &\textbf{Precision}  \\
\hline
\textbf{This is a picture about Picture of...}         &\ ViT   &\ 55.90    \\ 
\textbf{The action in the picture is... }              &\ ViT   &\ 56.58     \\ 
\textbf{A photo of a...}    &\ ViT   &\ 56.29      \\ 
\textbf{The content of the playing card is...  }       &\ ViT  &\ 57.64      \\ 
\textbf{None   }         &\ ViT   &\ 57.02     \\
\hline
\end{tabular}
\end{table}

\subsection{Visualization} \label{visualizations}

\begin{figure*}
    \centering
    \includegraphics[width=\linewidth]{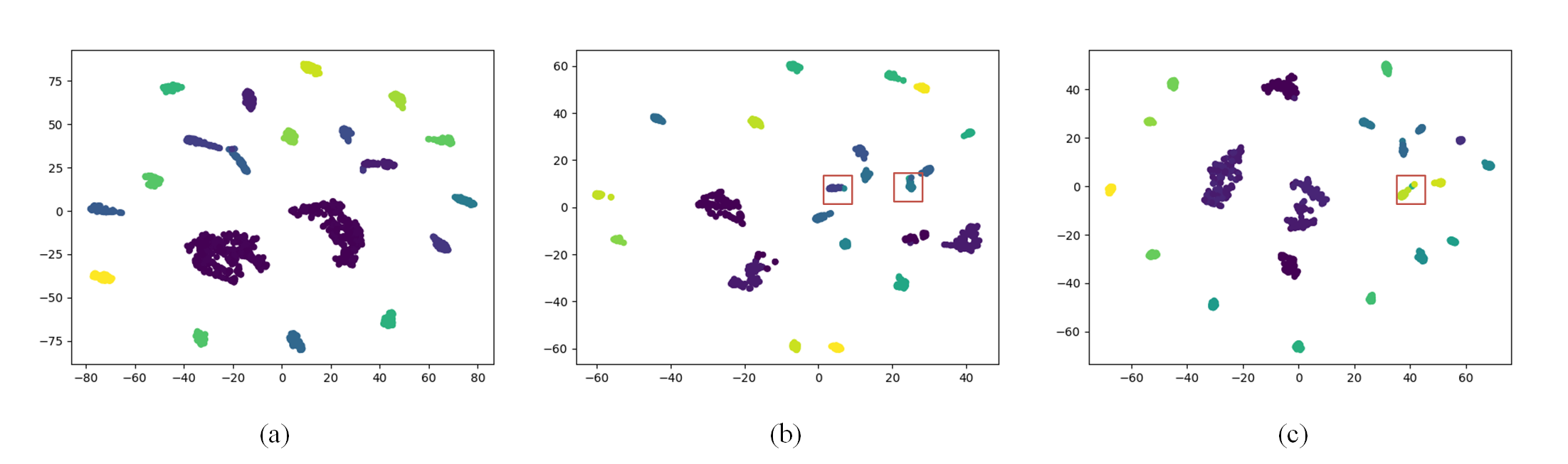}
    \caption{Visualization of feature distribution of (a) Ours, (b) VTB~\citep{62cec7b65aee126c0f5794bc}, and (c) VTF~\citep{zhu2023VTF} on PokerEvent.} 
    \label{fig:featEmbedding}
\end{figure*}

In this part, we give a quantitative analysis to further enhance the interpretability of our algorithm. The feature embedding and the top-5 recognition results are provided in the following subsections, respectively.

\noindent 
\textbf{Feature Embedding.~}  
As shown in Fig.~\ref{fig:featEmbedding}, we present a compelling visual example that allows us to delve into the distances between different classes. A total of 20 classes are randomly chosen to visualize this feature.Our study demonstrates that the performance of our proposed SAFE model shows a moderate improvement over the baseline model. Furthermore, when compared to VTF~\citep{zhu2023VTF}, a noticeable enhancement in performance can be observed. These visually illustrative representations underscore the capability of our proposed module in handling both RGB frames and event streams. The outcomes underline the incremental progress our model contributes, thereby opening avenues for potential opportunities in the field.

\begin{figure*}
    \centering
    \includegraphics[width=\linewidth]{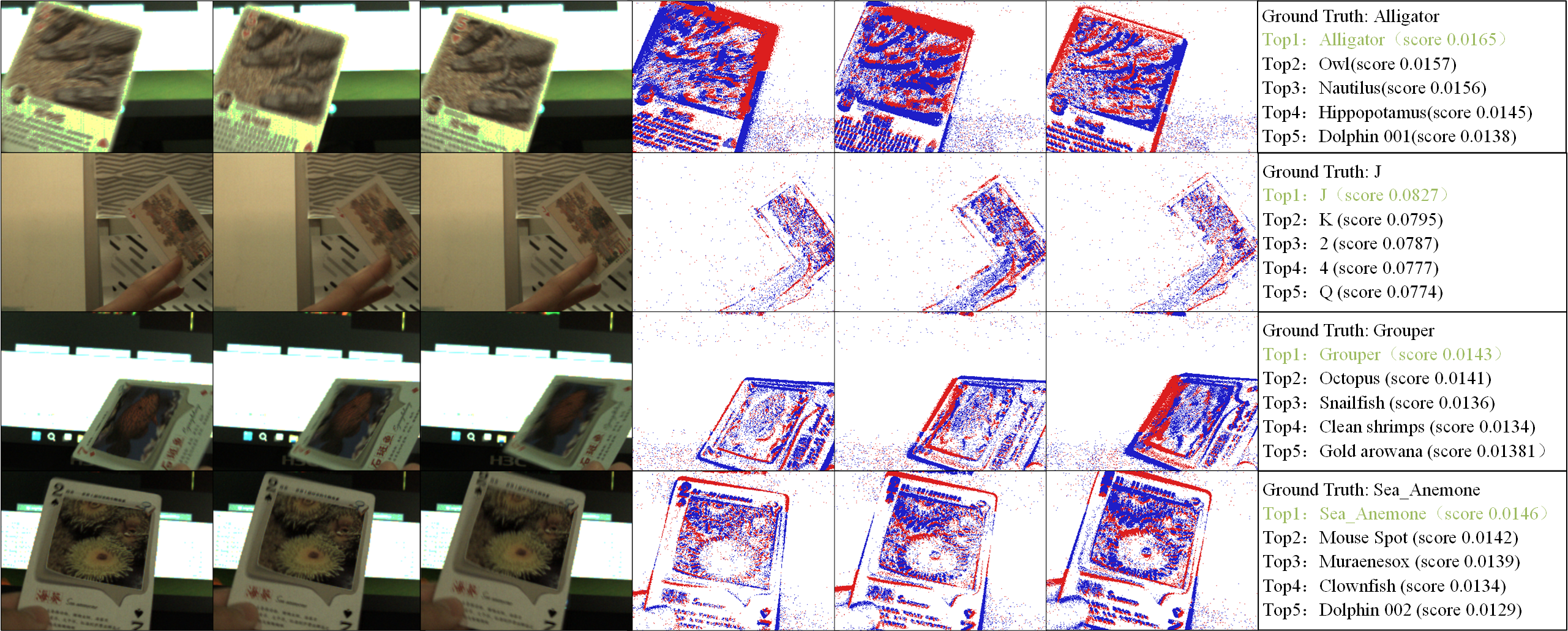}
    \caption{Visualization of the top-5 predicted results on the PokerEvent dataset.}
    \label{fig:poker}
\end{figure*}

\begin{figure*}
    \centering
    \includegraphics[width=\linewidth]{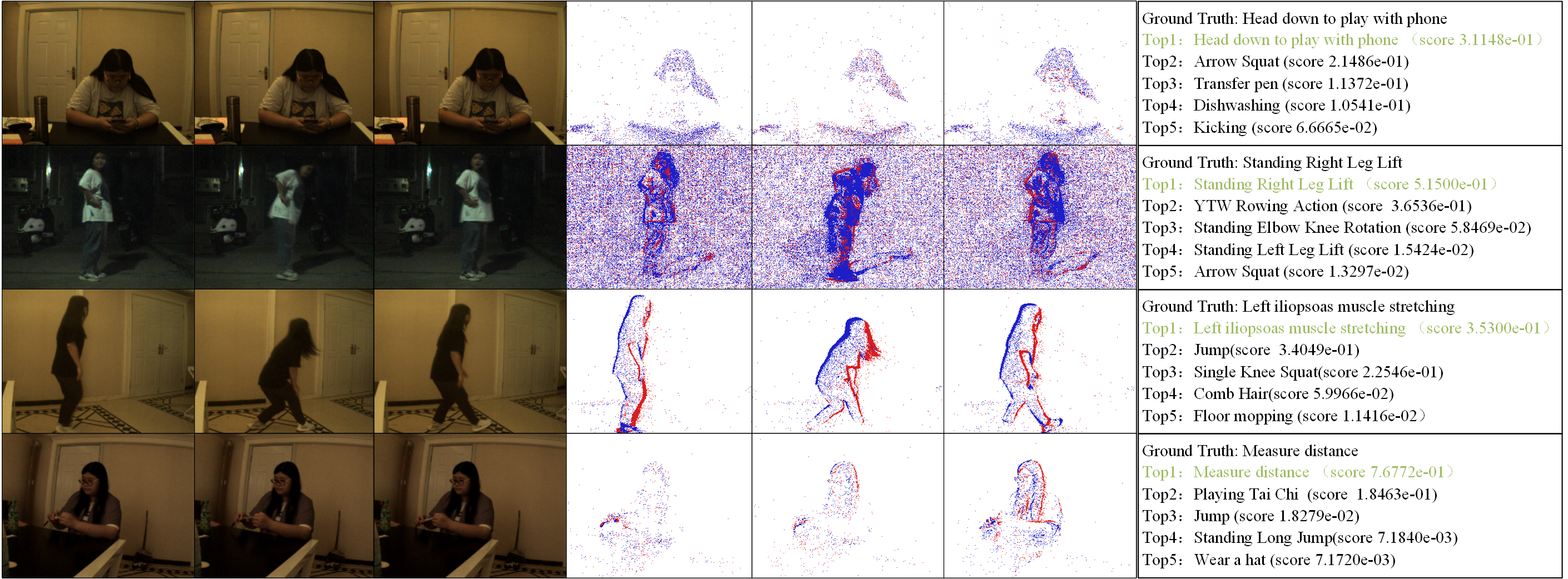}
    \caption{Visualization of the top-5 predicted results on the HARDVS dataset.}
    \label{fig:hardvs}
\end{figure*}

\noindent 
\textbf{Top-5 Recognition Results.~} 
As depicted in the Fig.~\ref{fig:poker} and Fig.~\ref{fig:hardvs}, we display four sets of RGB event samples from both the PokerEvent and HARDVS datasets, alongside their corresponding visualizations of the top five recognition results. These scores represent the probability scores for each of the 114 categories in the PokerEvent dataset and 300 categories in the HARDVS dataset, and the sum of these scores equals 1. From these images, we can observe that the RGB modality is susceptible to motion blur, while the event stream effectively filters out static background information and captures motion information proficiently, thereby enhancing target recognition.

\subsection{Limitation Analysis}   \label{limitations} 
Although significant progress has been made in the proposed SAFE model, there are still some limitations that need further exploration. Firstly, the SAFE model heavily relies on pre-trained large-scale vision-language models, such as CLIP, for feature extraction. Therefore, it may not be optimal for specific tasks or datasets with unique features. Secondly, computational resources can pose challenges for its deployment in real-time applications or on devices with limited computing capabilities. Given that the model involves complex feature extraction and fusion processes, it may require more computational power and storage capacity.

\section{Conclusion and Future Works} \label{conclusions}  

Current methods typically employ backbone networks to individually extract the features of RGB frames and event streams, and subsequently fuse these features for pattern recognition. However, we posit that these methods may suffer from two key issues: 
1). They attempt to directly learn a mapping from input vision modalities to the semantic labels. This approach often leads to sub-optimal results due to the disparity between the input and semantic labels; 
2). They utilize small-scale backbone networks for the extraction of RGB and Event input features, thus these models fail to harness the recent performance advancements of large-scale visual-language models. 
To address these issues, in this paper, we propose a novel pattern recognition framework that considers the semantic category information when fusing the RGB and Event data using a large vision-language model. Specifically, we employ a pre-trained large-scale vision model (CLIP vision encoder) to extract the RGB and event features. To handle the semantic labels, we initially convert them into language descriptions through prompt engineering and then obtain the semantic features using the pre-trained large-scale language model (CLIP text encoder). Subsequently, we integrate the RGB/Event features and semantic features using multi-modal Transformer networks. The resulting frame and event tokens are further amplified using self-attention layers. Concurrently, we propose to enhance the interactions between text tokens and RGB/Event tokens via cross-attention. Finally, we consolidate all three modalities using self-attention and feed-forward layers for recognition. Comprehensive experiments on the HARDVS and PokerEvent datasets fully substantiate the efficacy of our proposed SAFE model. 
In our future works, we will consider pre-training a large-scale RGB-Event big model to further improve the representation of event streams.


\bibliographystyle{apalike}
\bibliography{reference}


\end{document}